\pdfoutput=1

\ifdefined\REVIEW
    \documentclass[
        a4paper,
    ]{article}
    \usepackage{lineno, setspace}
    \modulolinenumbers[5]
    \pagewiselinenumbers
\else
    \documentclass[
        a4paper,
        twocolumn
    ]{article}
\fi

\usepackage{IEK10}

\usepackage{pifont}

\def\IEK10{
  Forschungszentrum Jülich GmbH,
  Institute of Energy and Climate Research,
  Energy Systems Engineering (IEK-10),
  Jülich 52428,
  Germany
}

\def\SVT{
  RWTH Aachen University,
  Process Systems Engineering (AVT.SVT),
  Aachen 52074,
  Germany
}

\def\Colone{
    Institute for Theoretical Physics, 
    University of Cologne, 
    50937 K\"oln, 
    Germany
}
\def\nmbu{Faculty of Science and Technology, Norwegian University of Life Sciences, 1432 Ås, Norway}

\newcommand{\mytitle}{
Multivariate Scenario Generation of Day-Ahead Electricity Prices using Normalizing Flows
}

\newcommand{\affil}{
  \begin{itemize}[leftmargin=3mm, itemsep=0mm]
    \item[$^a$]\IEK10
    \item[$^b$]\Colone
    \item[$^c$]\nmbu
    \item[$^d$]\SVT
  \end{itemize}
}

\def\firstAuthor{Hannes Hilger}
\newcommand{\myauthor}{
\firstAuthor$^{a,b}$\orcidlink{0009-0002-4772-0083},
Dirk Witthaut$^{a,b}$\orcidlink{0000-0002-3623-5341},
Manuel Dahmen$^{a}$\orcidlink{0000-0003-2757-5253},
Leonardo Rydin Gorj\~ao$^{c}$\orcidlink{0000-0001-5513-0580},
Julius Trebbien$^{a,b}$\orcidlink{0000-0001-9831-876X},
Eike Cramer$^{d,a,*}$\orcidlink{0000-0002-6882-5469} }
\newcommand{\correspondingauthor}{Eike Cramer, \SVT \newline E-mail: eike.cramer@alumni.tu-berlin.de}
\author{\myauthor}

\usepackage[
  colorlinks,
  linkcolor=blue,
  citecolor=blue,
  urlcolor=blue,
  pdftitle={\mytitle},
  pdfauthor={\firstAuthor}
]{hyperref}
\usepackage[capitalise, nameinlink]{cleveref}
\crefname{table}{Tab.}{Tab.}

\usepackage{orcidlink}

\newcommand{\setpgfexternalcounter}[1]{
  \makeatletter \pgfkeysgetvalue{/tikz/external/figure name}\myexternalname
  \expandafter\gdef\csname c@tikzext@no@\myexternalname\endcsname{#1}\makeatother
}

\begin{document}

\ifx\REVIEW\undefined
\twocolumn[
\begin{@twocolumnfalse}
\fi
  \thispagestyle{firststyle}

  \begin{center}
    \begin{large}
      \textbf{\mytitle}
    \end{large} \\
    \myauthor
  \end{center}

  \vspace{0.5cm}

  \begin{footnotesize}
    \affil
  \end{footnotesize}

  \vspace{0.5cm}

    \begin{abstract}
    Trading on the day-ahead electricity markets requires accurate information about the realization of electricity prices and the uncertainty attached to the predictions. 
    Deriving accurate forecasting models presents a difficult task due to the day-ahead price's non-stationarity resulting from changing market conditions, e.g., due to changes resulting from the energy crisis in 2021. 
    We present a probabilistic forecasting approach for day-ahead electricity prices using the fully data-driven deep generative model called normalizing flow. 
    Our modeling approach generates full-day scenarios of day-ahead electricity prices based on conditional features such as residual load forecasts. 
    Furthermore, we propose extended feature sets of prior realizations and a periodic retraining scheme that allows the normalizing flow to adapt to the changing conditions of modern electricity markets. 
    Our results highlight that the normalizing flow generates high-quality scenarios that reproduce the true price distribution and yield accurate forecasts. 
    Additionally, our analysis highlights how our improvements towards adaptations in changing regimes allow the normalizing flow to adapt to changing market conditions and enable continued sampling of high-quality day-ahead price scenarios.  
\end{abstract}


  \vspace*{5mm}
\ifx\REVIEW\undefined
\end{@twocolumnfalse}
]
\fi

  \section{Introduction}
Modern electricity markets such as the European Power Exchange (EPEX) support the transition to a more sustainable energy system.
Here, electricity is traded on short-term spot markets such as the day-ahead or the intraday market that provide structured trading intervals of either one hour or 15-minute blocks~\cite{european_network_of_transmission_system_operators_for_electricity_transparency_nodate}. 
Accurate anticipation of electricity prices on these markets allows consumers and producers to plan ahead to maximize their financial objectives and secure safe operation. 
Thus, electricity price forecasting is of central importance for energy system operation but remains challenging. 

Short-term markets like the day-ahead market depend on the demand and the generation from renewable electricity sources~\cite{trebbien2023understanding, wolff2017short}. 
Renewable electricity generation is intrinsically uncertain and fluctuates on various time scales from minutes to seasons~\cite{anvari2016short, staffell2018increasing}. 
Furthermore, electricity markets are non-stationary, i.e., they evolve in time due to changes in the generation mix, the regulatory framework, or geopolitical circumstances. 
For instance, the European electricity markets underwent a fundamental change in late 2021 caused by the energy crisis related to the war in Ukraine starting in February 2022, leading to exploding prices and substantial changes in the behavior of the electricity prices~\cite{goldthau2022energy, bottcher2023initial}. 
The distribution of prices is non-Gaussian with heavy-tails and occasional negative values, and price changes are strongly correlated over several hours~\cite{han2022complexity}. 
We argue that electricity price forecasting models must be able to adapt to changes while capturing the intrinsic uncertainty of the market by accurately describing the electricity price's probability distribution.

We present a data-driven, adaptable, and probabilistic forecasting model to generate scenarios of day-ahead electricity prices.
Our model learns the conditional distribution of day-ahead electricity prices based on forecasts of external factors such as wind and solar power generation and load.
We model all 24 hourly day-ahead prices for a given day as a multivariate joint probability distribution.
This multivariate probabilistic forecasting approach reflects the fundamental structure of the day-ahead electricity markets, where all 24 hourly prices are set simultaneously \cite{european_network_of_transmission_system_operators_for_electricity_transparency_nodate}. 
To learn the conditional probability distribution, we use conditional normalizing flows \cite{winkler2020learning, rasul2021multivariate}, which we previously used for wind power scenario generation~\cite{cramer2022normalizing} and prediction of intraday electricity prices~\cite{CRAMER2023_deltaID3}.
The conditional normalizing flow is a deep generative model~\cite{goodfellow2016deep} based on invertible neural networks~\cite{papamakarios2021normalizing}. 
Ensemble forecasting or scenario generation approaches, such as normalizing flows, provide several advantages over simpler methods like point forecasting or forecasting of mean and standard deviation \cite{cramer2022principal}. 
Scenario forecasts can produce potentially complicated, non-Gaussian forecast distributions. 
Moreover, each scenario is intrinsically consistent, i.e., correlations between the time steps are considered and reproduced. 
Additionally, the generated scenarios enable the formulation and solution of stochastic optimization problems to plan ahead under uncertainty~\cite{cramer2022normalizing, beykirch2022bidding}.

We design our model architecture to be robust to changes in the overall market behavior such as the price increase resulting from the energy crisis in 2021 and the ongoing war in Ukraine. 
The model inherits price, demand, and renewable power generation data from the previous day as conditional inputs. 
Thus, the model can rapidly detect changes and adapt accordingly. 
Furthermore, we propose a periodic model update through regular retraining steps. 
The retraining allows the model to compensate for fundamental changes in market structure and behavior such as regulatory changes or the increasing share of renewables. 

The model is trained and tested using data from the German-Luxembourg day-ahead electricity market and power system. We evaluate the model performance and provide a detailed statistical analysis, comparing predictions and the actual price time series. The results show that the model reproduces the intricate statistical properties of the price time series, including the heavy-tailed distribution as well as conditional distributions, temporal correlations, and the impact of the European energy crisis.

The article is organized as follows: We first provide some background on the European electricity markets and review the state of the art in electricity price forecasting in Section~\ref{sec:background}. Then, we describe the concept and implementation of the normalizing flow in Section~\ref{sec:methods}. Our results on the model performance and the statistical properties of prices and scenarios are given in Section~\ref{sec:results}. Finally, we summarize and discuss our results in Section~\ref{sec:conclusion}.

\section{Background}
\label{sec:background}

This Section reviews the structural setup of the European electricity markets including the day-ahead bidding markets. In the second part of the Section, we review the state-of-the-art in electricity price forecasting. 

\subsection{European electricity markets}
\label{sec:markets}

Stable operation of an electric power system requires that power generation and demand are balanced at all times~\cite{kundur2007power}. In the European system, power generation is mainly coordinated through trading on electricity markets on different time scales, e.g., in hourly or quarter-hourly intervals. 
Each market participant has to align the physical net amount of electrical energy that is produced or consumed in a given time window to the ``virtual'' amount of electrical energy that is bought or sold on the electricity markets in that particular time window~\cite{bundesnetzagentur_bundesnetzagentur_nodate}. For instance, a wind farm operator is required to market the exact amount of electricity produced in any given quarter-hourly time window. This process ensures a physical balance between power generation and demand on the system level. Residual imbalances lead to deviations of the grid frequency from its set value of 50\,Hz and are corrected in real-time via the load-frequency control systems~\cite{kundur2007power, bundesnetzagentur_bundesnetzagentur_nodate}. Generally, the daily and weekly patterns of buy and sell decisions lead to complex fluctuations of electricity prices~\cite{han2022complexity}. 

Market participants may buy and sell electricity either via direct power purchase agreements, which may be agreed on months or years in advance, or via trading on an electricity exchange~\cite{bundesnetzagentur_bundesnetzagentur_nodate}. On the exchanges, electricity is traded on the futures markets and the spot markets. Power futures are long-term contracts that regard delivery dates months or years in advance. On the spot markets, electricity is traded with delivery dates on the next day (day-ahead)~\cite{huisman2007hourly} or the same day (intraday)~\cite{shinde2019literature}.

Trading is organized in bidding zones and we will focus on the Germany-Luxembourg bidding zone (Germany-Austria-Luxembourg until October 1, 2018). For this article, we will restrict our analysis to the European Power Exchange EPEX Spot \cite{european_network_of_transmission_system_operators_for_electricity_transparency_nodate}, which has the highest trading volume for the Germany-Luxembourg bidding zone. 
Furthermore, we focus on the day-ahead market, the most important spot market in terms of trading volume~\cite{huisman2007hourly}. At EPEX Spot, electricity is traded in hourly windows for the 24 hours of the following day. Market participants place buy and sell orders until 12:00. Then, the hourly prices are determined according to the market clearing principle: The highest price that finds a buyer in each hour is determined as the market clearing price for that hour. Every unit of electricity is traded at the market clearing price in each respective hour. This is commonly referred to as ``pay-as-cleared''. Predicting this market clearing price is the central objective of this article.

\begin{figure}[t]
\centering
\includegraphics[width=\columnwidth]{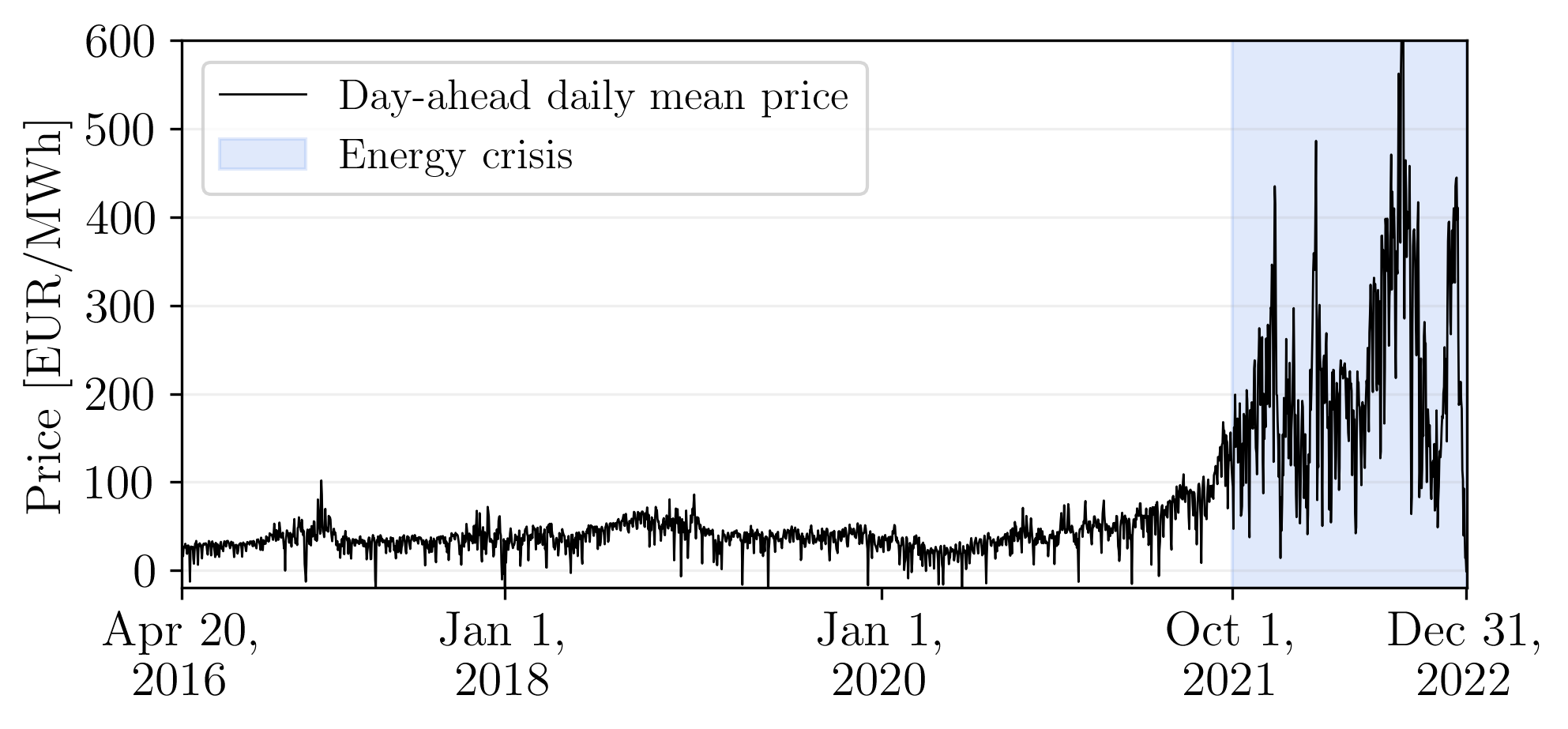}
    \caption{Time series of day-ahead mean prices of each day from April 20, 2016 to December 31, 2022. We consider October 1, 2021, as the beginning of the 2021/22 energy crisis (shaded period). Data from EPEX Spot, taken from the ENTSO-E transparency platform \cite{european_network_of_transmission_system_operators_for_electricity_entso-e_nodate}.
    }
    \label{fig:1}
\end{figure}
The European electricity markets were strongly affected by the energy crisis of 2021 and 2022 related to the ongoing war in Ukraine. Energy prices soared in many regions of the world in 2021~\cite{goldthau2022energy}. Europe was particularly strongly affected, as many countries were dependent on fossil fuel imports from the Russian Federation. 
Figure~\ref{fig:1} shows the daily average day-ahead prices in the Germany-Luxembourg bidding zone from April 20, 2016, to December 31, 2022. The average price level soared from around 30\,EUR/MWh before the crisis to around 200\,EUR/MWh during the crisis, with peaks up to 800\,EUR/MWh. Notably, the energy crisis began well before the beginning of the war in Ukraine in late February 2022 due to rising political tensions in the preceding months. As the beginning of the energy crisis is not clearly defined, we use October 1, 2021, as a reference date during our analysis.

\subsection{Electricity price forecasting and scenario generation}
\label{sec:EPF}

The field of electricity price forecasting is well established and receives contributions from economics and technical fields like engineering, computer science, and physics \cite{WERON2014_review}. 
There are works concerned with day-ahead electricity prices \cite{LAGO2021116983, wolff2017short} as well as intraday electricity prices, e.g., our previous work on normalizing flows \cite{CRAMER2023_deltaID3}.

Traditionally, electricity price forecasting relied on statistical time series models such as autoregressive (ARIMA, LASSO) models \cite{WERON2014_review}. 
However, with the increase in computing power and research on neural network regression, deep learning became one of the drivers for continuous development in electricity price forecasting \cite{jkedrzejewski2022electricity}. Here, artificial neural networks and time series neural networks like Long-Short Term Memory (LSTM) models are the workhorse methods \cite{KAPOOR2023121446, trebbien2023probabilistic}. 
Despite the increased understanding of modern electricity markets, the realization of day-ahead electricity prices remains a stochastic process. 
Thus, measures of uncertainty such as probabilistic forecasts can greatly improve the reliability of the predictions \cite{weron2018_Probabilistic}. 
Other approaches to quantify the uncertainty include ensemble forecasts~\cite{narajewski2020ensemble}, generation of prediction intervals for neural network forecasts~\cite{KHOSRAVI2013120}, moment matching~\cite{weron2018_Probabilistic}, or quantile regression~\cite{uniejewski2021regularized}. Other works use combinations of deterministic and probabilistic forecasting to balance between accurate forecasting and uncertainty quantification \cite{marcjasz2020probabilistic}. 
Recently, probabilistic forecasting also relies on machine learning instead of established statistical modeling. For instance, Xu~et~al.~\cite{Xu2024quantileregression} propose a deep learning scheme for quantile regression based on kernel density estimation. Other works also rely on deep learning, e.g., by using ensemble forecasting via time series regression models like LSTM models~\cite{Bozlak2024deeplearning}.
Marcjasz~et~al.~\cite{Marcjasz2023distributionalANN} use distributional neural networks to predict full distributions. The distributional neural network predicts the parameters of predefined distribution models such as Gaussian or Gamma distributions. Their study shows the unbounded Johnson's $S_U$ distribution to be the most accurate approximation for day-ahead prices among their trials.  

Most of the published approaches to forecasting day-ahead electricity prices rely on a step-by-step forecasting approach, e.g., in autoregressive models \cite{LAGO2021116983, Xu2024quantileregression, Bozlak2024deeplearning}. 
Notably, such an approach contrasts the actual procedure of settling the day-ahead bidding markets, where all 24 hourly price intervals are set simultaneously (cf.~Section~\ref{sec:markets}). 
Instead, multivariate forecasting matches the fundamental structure of the day-ahead market. Ziel and Weron~\cite{ziel2018day} compare univariate and multivariate forecasting and report improved performance for the multivariate case. Other works combine multivariate forecasting with Schaake shuffles to obtain probabilistic methods~\cite{GROTHE2023_multi_probabilistic}. Klein~et~al.~\cite{Klein2023copulaANN} use copula methods in combination with deep neural networks for forecasting intraday prices in the Australian market. 
Our previous work~\cite{CRAMER2023_deltaID3} is the only work using normalizing flows to predict electricity prices. 
In contrast to the present paper, our previous work considers the problem of intraday price forecasting. 

The multivariate full-day scenario generation approach using a deep generative model we implement in this work has precedent in renewable power generation scenarios. 
For instance, Chen~et~al.~\cite{chen2018model} use generative adversarial networks (GANs) to generate scenarios of photovoltaic and wind power generation.
Qi~et~al.~\cite{qi2020optimal} use variational autoencoders (VAEs) to generate scenarios of concentrated solar power for optimization of multi-energy systems. 
Both GANs and VAEs are powerful generative models, however, they are dependent on unreliable training schemes that are not guaranteed to yield adequate results. 
Normalizing flows are trained using direct log-likelihood maximization, which yields numerically consistent results \cite{papamakarios2021normalizing}.
In our previous works \cite{cramer2022principal, cramer2022normalizing}, we have compared the normalizing flow with GANs and have found the normalizing flow to yield superior results in all considered metrics.

Table~\ref{tab:1} lists a comparison of methods used for scenario generation and electricity price forecasting. 
Note that only the normalizing flow combines full-day scenario generation with non-Gaussian statistics and a reliable training method. 
\begin{table*}[]
\centering
\caption{Comparison of methods for electricity price forecasting and scenario generation.}
\resizebox{\textwidth}{!}{\begin{tabular}{@{}lccccc@{}}
\toprule
                        & \makecell{Reliable\\training} & Day-specific & \makecell{Consistent with\\market structure} & \makecell{Uncertainty \\quantification} & \makecell{Non-Gaussian\\ statistics} \\ \midrule
Autoregressive models \cite{LAGO2021116983, wolff2017short}   & \ding{51}         & \ding{51}    & \ding{55}                        & \ding{55}                  & \ding{55}               \\
Moment matching \cite{weron2018_Probabilistic}         & \ding{51}         & \ding{51}    & \ding{55}                        & \ding{51}                  & \ding{55}               \\
Moment forecasting \cite{Klein2023copulaANN}     & \ding{51}         & \ding{51}    & \ding{55}                        & \ding{51}                  & \ding{55}               \\
Multivariate regression \cite{ziel2018day} & \ding{51}         & \ding{51}    & \ding{51}                        & \ding{55}                  & \ding{55}               \\
GANs \cite{chen2018model} and VAEs \cite{qi2020optimal}           & \ding{55}         & \ding{51}    & \ding{51}                        & \ding{51}                  & \ding{51}               \\
Normalizing Flow (our)  & \ding{51}         & \ding{51}    & \ding{51}                        & \ding{51}                  & \ding{51}               \\ \bottomrule
\end{tabular}}
\label{tab:1}
\end{table*}

There are a few works considering adaptations towards changing market conditions, although the importance of adaption became obvious during the energy crisis. 
Examples include adaptive preprocessing~\cite{sebastian2023adaptive} and our previous work on probabilistic forecasting using LSTM models~\cite{trebbien2023probabilistic}. 
Please note that our previous work on normalizing flow-based intraday electricity price forecasting does not consider any adjustment to changing market conditions. 

Recent advances in machine learning have benefited both model development as well as feature selection for forecasting. 
For instance, our previous work uses SHapley Additive exPlanations (SHAP) values to dissect the functional relationship between electricity prices and relevant features beyond the merit order principle \cite{trebbien2023understanding}.
In a similar work, Tschora~et~al.~\cite{TSCHORA2022118752} use SHAP values to identify correlations between bidding zones to improve their forecasting performance. 

\section{Methods and Data}
\label{sec:methods}

\subsection{Fundamentals of normalizing flows}

Normalizing flows are a class of deep generative models using invertible transformations. 
The concept of normalizing flows was first introduced by Tabak and Vanden-Eijnden~\cite{tabak2010density} and Tabak and Turner~\cite{tabak2013family} about ten years ago. 
A generative model describes the probability distribution of a given data set and can generate new samples from that distribution. 
Notably, other generative models like VAEs \cite{kingma2014auto} and GANs \cite{goodfellow2014generative} give an implicit representation of the probability distribution, i.e., they only allow for sampling. 
Normalizing flows, however, provide an explicit representation of the probability distribution, i.e., the probability density function (PDF), which enables mathematically consistent and efficient training via likelihood maximization. 
We refer to Papamakarios~et~al.~\cite{papamakarios2021normalizing} for a comprehensive review of normalizing flows.

The target data, in our case the day-ahead electricity prices, is represented by a random vector $X \in \mathbb{R}^{D}$.
The model learns a diffeomorphism~\cite{papamakarios2021normalizing}, i.e., a differentiable invertible transformation \begin{align*}
    f: \; & \mathbb{R}^{D}  \rightarrow \mathbb{R}^{D} \\
       & x \mapsto f(x) 
\end{align*}
that maps $X$ to another random variable $Z$ following a well-known base distribution.
The most common choice for the base distribution is a multivariate standard normal (Gaussian) distribution, i.e., $Z \sim N(\mathbf{0}, \mathbf{I})$. Using the diffeomorphism, normalizing flows provide an explicit representation of the PDF of the target variable $X$ via a change of variables~\cite{papamakarios2021normalizing}, i.e., 
\begin{align}\label{eq:1}
    p_X(x) = p_Z(f(x))  \left|  \det J_f(x) \right|^{-1},
\end{align}
where $J_f(x)$ denotes the Jacobian of the function $f$ at the point $x$.
This direct representation allows for sampling according to $p_X(x)$ by first sampling $z$ from the Gaussian $p_Z(z)$ and then transforming it through the inverse transformation, i.e., computing $x = f^{-1}(z)$.

Using the explicit PDF in Equation~\eqref{eq:1}, a normalizing flow is trained via likelihood maximization~\cite{papamakarios2021normalizing}. 
Let $x_1, x_2, \ldots, x_N$ denote the data points from the respective training set.
Then, the function $f$ is chosen such that it minimizes the negative log-likelihood
\begin{align}
    NLL = - \sum_{i=1}^N \log \left[ p_Z\left(f(x_i)\right)  \left|  \det J_f(x_i) \right|^{-1}   \right].
\end{align}
In practice, $f$ is chosen as an invertible neural network with a finite set of parameters $\theta$.

\begin{figure}\centering
    \includegraphics[width=\columnwidth]{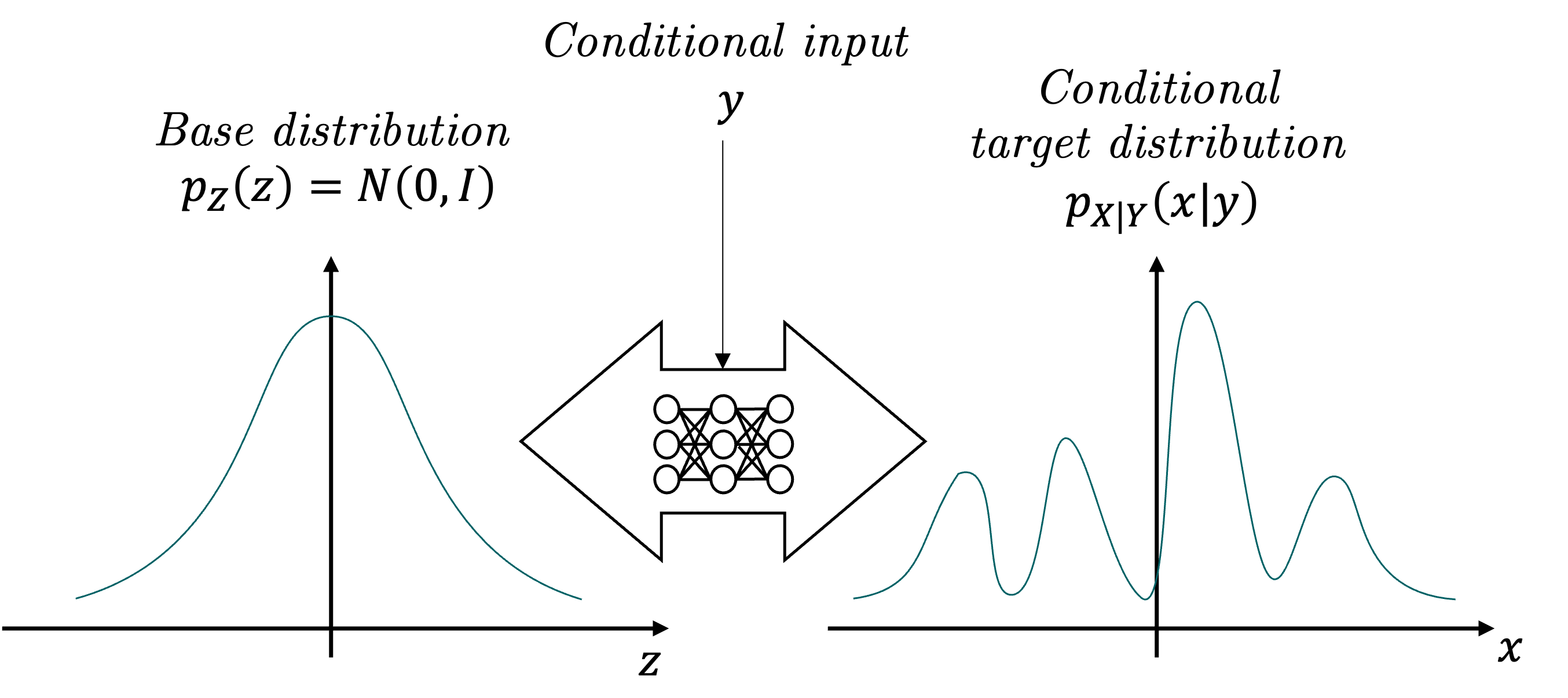}
    \caption{Schematic visualization of the conditional normalizing flow model with presentation of one-dimensional probability density functions. The left side represents the known base distribution $p_Z(z)$. The right side represents the conditional non-Gaussian target distribution $p_{X|Y}(x|y)$. The network in the center shows the diffeomorphism in Equation~\eqref{eq:3} between the two distributions, which depends on a conditional input $y$.}
    \label{fig:2}
\end{figure}

The baseline normalizing flow can be extended to conditional statistics \cite{rasul2021multivariate, cramer2022normalizing}, where the probability distribution depends on another variable $y \in \mathbb{R}^{L}$.
This conditional input is taken into account by generalizing the flow to
\begin{equation}\label{eq:3}
\begin{aligned}
    f: \; & \mathbb{R}^{D} \times \mathbb{R}^{L} \rightarrow \mathbb{R}^{D} \\
       & x, y \mapsto f(x, y). 
\end{aligned}
\end{equation}
For every fixed value of $y$, the restricted function $x \mapsto f(x, y)$ must be differentiable and invertible~\cite{winkler2020learning} w.r.t.~$x$.
Then, the conditional PDF is given by
\begin{align}
    p_{X|Y}(x|y) = p_Z\left(f(x,y)\right) \left|  \det J_f(x,y) \right|^{-1},
\end{align}
where $J_f(x,y)$ denotes the Jacobian with respect to the variable $x$.
Figure~\ref{fig:2} shows a schematic visualization of the conditional normalizing flow including the standard normal base distribution and the conditional non-Gaussian target distribution. 
The conditional inputs are considered as additional input to the diffeomorphism. 

The extension to conditional distributions allows us to use the normalizing flow as a multivariate probabilistic regression model.
This is not restricted to a particular probability distribution \cite{winkler2020learning, cramer2022normalizing}.
If the diffeomorphism is constructed using flexible functions such as neural networks, the normalizing flow becomes highly expressive and can describe any type of conditional distribution \cite{papamakarios2021normalizing}. Furthermore, the use of neural networks and training alleviates the need to make special considerations of correlations and interdependencies of the conditional inputs.
The fitting of normalizing flows automatically learns such dependencies and considers them in the later scenario generation \cite{winkler2020learning, cramer2022normalizing}.

To sample scenarios using the normalizing flow, we sample random instances $\hat{z}$ from the Gaussian distribution $Z \sim N(\mathbf{0}, \mathbf{I})$ and transform these instances using the inverse of $f$:
\begin{equation}
    \hat{x} = f^{-1}(\hat{z},y)
\end{equation}
Here, $\hat{x}$ are the generated scenarios based on the conditional inputs $y$.

\subsection{Model architecture and training}
\label{sec:architecture}

We implement the conditional normalizing flow using the real non-volume preserving transformation (RealNVP)~\cite{dinh2016density} with an extension to include conditional features \cite{winkler2020learning, cramer2022normalizing}.
RealNVP uses affine coupling layers that construct highly flexible transformations that guarantee the invertibility of the overall transformation.
The coupling layers are built on so-called conditioner models that introduce nonlinearity into the transformation. 
For more details on normalizing flows and their implementation, we refer to the review article by Papamakarios~et~al.~\cite{papamakarios2021normalizing}, the original work on RealNVP by Dinh~et~al.~\cite{dinh2016density}, and our previous works~\cite{cramer2022principal, cramer2022normalizing}.

As conditional inputs, we use the concatenation of seven 24-dimensional forecast profiles, which amounts to a 168-dimensional conditional input vector $y$ that is passed to the conditional RealNVP layers. The conditional inputs include the day-ahead forecasts of wind and solar generation and load for every hour of the following day as these features show the highest influence on the realization of the day-ahead prices~\cite{trebbien2023understanding}.
Furthermore, the conditional input also includes the wind, solar, and load forecasts and the day-ahead price realization of the previous day.
The latter information allows the model to scale the predicted day-ahead prices.

We rely on the publicly available data in the ENTSO-E transparency platform~\cite{european_network_of_transmission_system_operators_for_electricity_entso-e_nodate}.
The ENTSO-E platform provides historical data and day-ahead forecasts of the residual load constituents.
We outline the full data preprocessing below. There is no hidden assumption about the availability of particular data or third-party forecasting models. 

Recall that the same wind, solar, and load vectors result in very different day-ahead prices before and during the energy crisis~\cite{trebbien2023probabilistic}.
Therefore, any model that is trained prior to and deployed during the energy crisis is likely to perform poorly.
Including information from the previous day solves this problem for two reasons.
First, it provides a typical price level for the respective period.
Second, the model can learn that a certain set of wind, solar, and load profiles resulted in a certain day-ahead price profile on the previous day.
Including this additional information enables the model to predict what the wind, solar, and load forecasts for the next day might result in.
The robustness of the model performance is assessed in detail in Section~\ref{sec:results}.

We scale all power data, i.e., the wind, solar, and demand data, by a factor of 1.1 times their historical maximum to obtain features between 0 and 1.
All price data is scaled by a constant factor of 100.
Note that normalizing flows are not restricted to any specific interval, but scaling the data typically improves their performance \cite{papamakarios2021normalizing}.

In the final stage, the model contains a decoding step that reduces the dimensionality of the day-ahead electricity price data.
By this step, we mitigate a problem that repeatedly occurs in energy time series forecasting: 
The strong correlation of time steps means that the target data $X$ lies on a lower-dimensional manifold in the target space $\mathbb{R}^{D}$ \cite{cramer2022principal}.
In such a case, normalizing flows typically learn smeared-out distributions~\cite{brehmer2020flows} and generate noisy scenarios \cite{cramer2022principal}. We mitigate this problem by dimensionality reduction to a lower dimensional space using principal component analysis (PCA)~\cite{goodfellow2016deep,cramer2022principal}.
That is, we encode an original data point $x$ according to $x' := U^\top (x - \bar x)$, where $U$ is the matrix of principal components, $\bar x$ is the sample mean, and $\top$ denotes the transpose.
The normalizing flow is trained on the encoded data $x' \in \mathbb{R}^{D'}$, and scenarios are decoded using the inverse of $U^\top$, i.e., $x := \bar x + U \, x'$.
In practice, we use an encoding into $D' = 14$ dimensions, which explains $99.5\%$ of the variance of the original data.

To test the performance of the normalizing flow, we do not use a fixed train-test-split but implement a retraining scheme:
Every 90 days, the normalizing flow is newly trained on all available data until that point.
For instance, the normalizing flow might be newly trained at the end of 2018 with all data available until then (Jan 2016 - Dec 2018).
This retraining also includes adjustments of the scaling factors for preprocessing, if necessary.
The newly trained normalizing flow is then used for scenario generation for the following 90 days. For instance, at the beginning of April 2019, the normalizing flow is then retrained again with all available data (Jan 2016 - Mar 2019). 
It is this retraining scheme that allows the model to take into account non-stationary market conditions, as the normalizing flow regularly gains new training samples that might exhibit novel market behavior.
Note that the 90-day retraining interval is a heuristic that proved to work well in preliminary tests. 

\subsection{Implementation and hyperparameter optimization}

The normalizing flow is implemented in Python 3.9.13 using TensorFlow 2.10.0~\cite{abadi2016tensorflow}. The code for the normalizing flow is based on our prior studies~\cite{cramer2022normalizing} and open source libraries from~\cite{team_keras_nodate}. The PCA calculations are done using scikit-learn 1.1.2 for Python~\cite{pedregosa2011scikit}.

\begin{table*}\centering
\caption{Hyperparameter optimization of the normalizing flow. We test different combinations of hyperparameters in two steps and evaluate the performance in terms of the mean absolute error (MAE).}
\label{tab:2}
\begin{tabular}{lcc}
\toprule
hyperparameter  & values (1st step)    & values (2nd step)  \\ \midrule
coupling layers & 2, 3, 4, 5           & 3, 4, 5 \\
network depth   & 2, 3, 4, 5           & 2, 3, 4 \\
network width   & 14, 21, 28           & 14, 21  \\
epochs          & 500, 750, 1000, 1500 & 1000\\
\bottomrule
\end{tabular}
\end{table*}

We use fully connected neural networks to implement the conditioner models. Thus, the model contains four hyperparameters: the number of coupling layers, the depth of each network describing the conditioner models, the number of nodes in each hidden layer of the conditioner models, and the number of training epochs. These hyperparameters are optimized in two steps using the JURECA DC supercomputer at Forschungszentrum Jülich~\cite{JURECA}.

For hyperparameter optimization, we follow the proposed retraining scheme from Section~\ref{sec:architecture} for all available data. Hence, the test data for each iteration are the 90 days following the latest cut-off. 
First, we train one model instance in the retraining scheme for each of the 192 different hyperparameter combinations as listed in the center of Table~\ref{tab:2}. 
In each case, we evaluate the mean absolute error (MAE) of the scenario mean and discard hyperparameter values that lead to high MAE values. 
We find that normalizing flows with just two coupling layers tend to underfit the data and thus discard this configuration.
In the second step, we train each parameter combination eight times and evaluate the mean and the standard deviation of the MAE to avoid an influence from stochastic effects in the training. 
Therefore, we reduced the number of parameter combinations according to the results of the first step, keeping only 18 combinations.
We list the six best-performing hyperparameter combinations in Table~\ref{tab:3}. 

\begin{table*}[t]
\centering
\caption{Results of the second step of hyperparameter optimization. We only show the six hyperparameter combinations with the lowest averaged MAE. For each hyperparameter combination, we train eight models and report the mean and the standard deviation over the eight runs.  
}
\label{tab:3}
\begin{tabular}{lcccc}
\toprule
 & \begin{tabular}[c]{@{}c@{}}coupling\\ layers\end{tabular} & \begin{tabular}[c]{@{}c@{}}hidden \\ layers\end{tabular} & \begin{tabular}[c]{@{}c@{}}hidden \\ nodes\end{tabular} & \begin{tabular}[c]{@{}c@{}}mean absolute error\\ {[}EUR/MWh{]}\end{tabular} \\  \midrule
 & 5 & 2 & 21 & $\textbf{11.11} \pm \textbf{0.56}$ \\
 & 4 & 2 & 14 & $11.17 \pm 0.24$ \\
 & 3 & 2 & 21 & $11.21 \pm 0.34$ \\
 & 3 & 3 & 21 & $11.29 \pm 0.20$ \\
 & 4 & 3 & 21 & $11.31 \pm 0.28$ \\
 & 4 & 2 & 21 & $11.32 \pm 0.21$ \\
 \bottomrule
\end{tabular}
\end{table*}

We find that the differences in performance between the different models are small and therefore the choice of hyperparameters appears to only play a minor role in the examined ranges. In the following, we choose the best-performing hyperparameter combination w.r.t.~the MAE (coupling layers: 5, number of hidden layers: 2, number of nodes: 21, epochs: 1000).

\subsection{Benchmark models}
\label{sec:benchmark}

To assess the performance of the normalizing flow, we consider two benchmark models for scenario generation. Similar to the normalizing flow, both benchmarks select full-day scenarios, i.e., electricity price time series covering the 24-hour day-ahead trading horizon. 
First, an uninformed historical model generates samples by randomly drawing from the pool of past full day-ahead price realizations.
For instance, on January 1, 2020, each scenario from the uninformed historical model is a price profile realization drawn randomly from the pool of price realizations from January 1, 2016, to December 31, 2019.
For each day, 50 scenarios are selected by randomly drawing 50 past price realizations.
The model ignores all conditional inputs but captures typical daily profiles.
We include this model to represent a valid reference point and lower bound for the model performance examination.

Second, an informed historical model generates samples using a k-nearest-neighbors approach.
It generates scenarios by drawing the historical price realizations of days with the closest conditional inputs w.r.t.~the Euclidean distance.
In other words, the generated scenarios consist of price profile realizations of the historical days with the most similar conditions.
The conditional vectors are a 96-dimensional concatenation of wind, solar, and load forecasts and the price realization of the previous day.
For each day, 50 scenarios are generated by determining the 50 days with the most similar conditions from the pool of past realizations and using the price profiles of these days as scenarios.
The k-nearest-neighbors model is implemented using the NearestNeighbors function from scikit-learn 1.1.2 in Python~\cite{pedregosa2011scikit}.

\subsection{Data Sources}

We use data from the ENTSO-E transparency platform \cite{european_network_of_transmission_system_operators_for_electricity_entso-e_nodate} from January 2016 to December 2022, which were retrieved via the restful API provided by ENTSO-E \cite{european_network_of_transmission_system_operators_for_electricity_transparency_nodate} using the entsoe-py open-source implementation for Python. 
The day-ahead price is the price of the EPEX Spot day-ahead auction, for the Germany-Luxembourg bidding zone (Germany-Austria-Luxembourg prior to October 1st, 2018).
The day-ahead load forecast is the expected hourly load in the Germany-Luxembourg bidding zone.
The day-ahead solar and wind forecasts are the expected hourly production of each generation type in the Germany-Luxembourg bidding zone. 
We use the ENTSO-E forecasts because they provide a coherent publicly available reference data source, although market participants typically use a variety of different forecasting products (cf.~the discussion in \cite{trebbien2023thesis}).

\section{Results}
\label{sec:results}
This Section analyzes the normalizing flow-generated scenarios of the day-ahead electricity price in comparison to the benchmark models.
We start by showing examples of normalizing flow-generated scenarios.
Next, we perform a statistical analysis of the generated scenarios and, finally, evaluate the scenarios using ensemble forecasting scores such as the energy score and the variogram score.

\begin{figure}\centering
\includegraphics[width=\columnwidth]{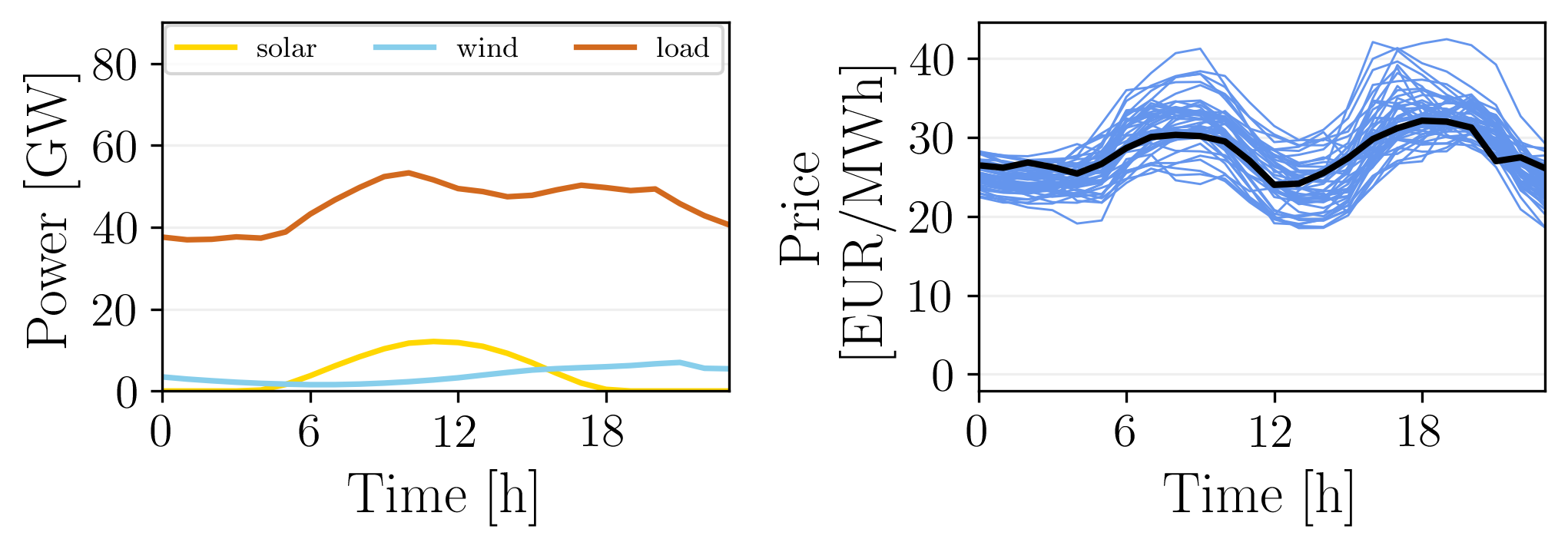}
    \includegraphics[width=\columnwidth]{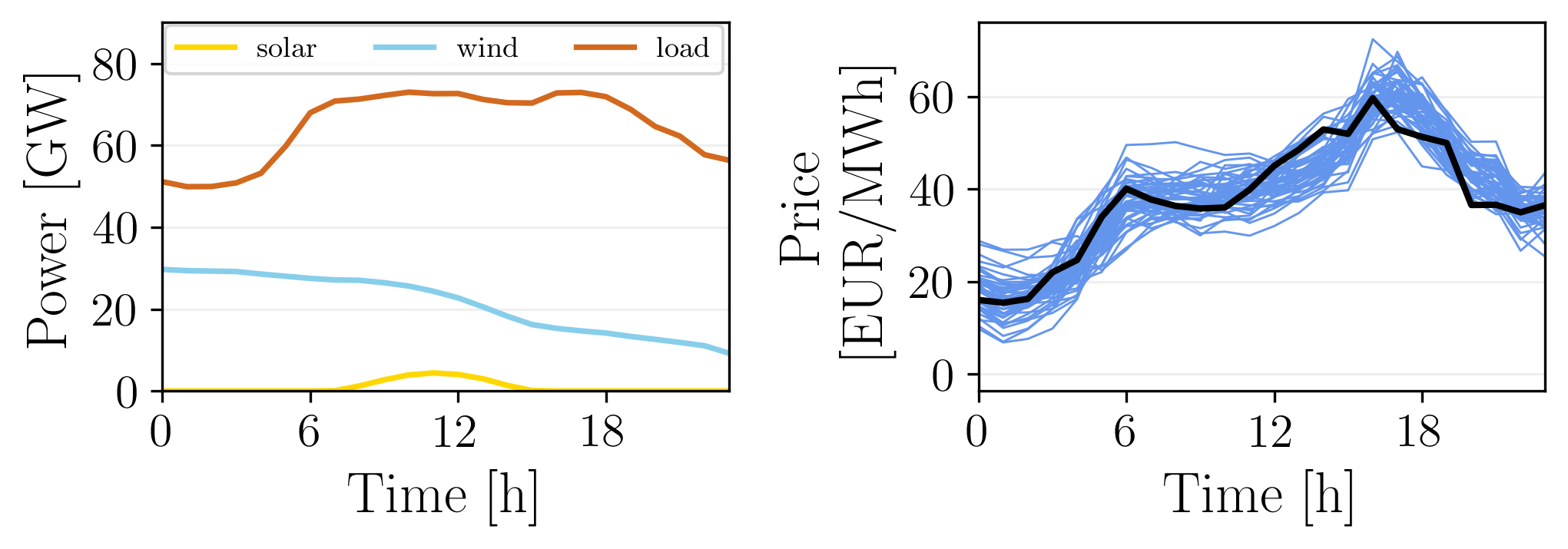}
    \includegraphics[width=\columnwidth]{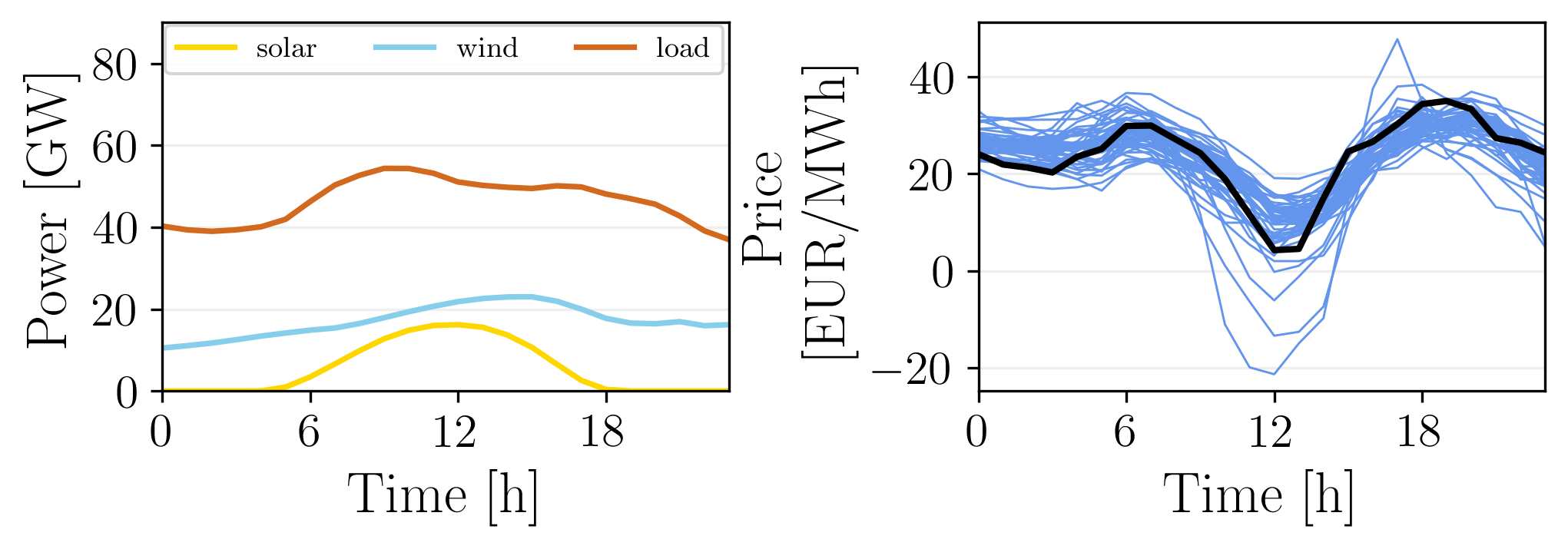}
    \caption{Example forecasts for May 7, 2017 (top), November 28, 2017 (center) and August 22, 2020 (bottom). The left column shows the solar generation forecast (yellow), wind generation forecast (blue), and load forecast (red) for each day. The right column shows 50 generated scenarios (blue) according to the conditions forecasts and respective price realization (black) for comparison.}
    \label{fig:3}
\end{figure}

\subsection{Initial examples}
This Section provides a qualitative overview of the capabilities of the conditional normalizing flow. Figure~\ref{fig:3} shows three selected examples of ensemble forecasts and the associated conditional inputs. The first row of Figure \ref{fig:3} shows a typical day in May 2017.
The load profile and the production from solar and wind of that day is on a low level, which translates into a typical price profile with two peaks, one in the morning and another in the afternoon. Prices are lower at noon due to stronger solar power generation and during the night due to a lower load. Overall, the shape and price level of the realization are well predicted by the scenarios from the normalizing flow.
The second row in Figure \ref{fig:3} shows a day where the expected wind power production is high in the morning hours but decreases throughout the day. This is well reflected in the day-ahead price profile scenarios, where the price peak in the afternoon is higher due to a higher residual load compared to that in the morning hours. Again, the generated scenarios tightly mirror the actual realization.
The third row in Figure \ref{fig:3} shows a day where the expected load is low (a typical Saturday), and the solar and wind productions are expected to be quite high, especially during the noon hours. Around noon, this combination results in a deep price dip in the day-ahead price to almost 0\,EUR/MWh. The model predicts this price dip and some scenarios even reach the negative price range. Here, the predicted price distribution becomes strongly non-Gaussian with a clear negative skewness.

\subsection{Statistical verification of normalizing flow-generated scenarios}
\label{sec:stats_nfresults}

Electricity price time series have intricate statistical properties~\cite{han2022complexity}, e.g., heavy-tailed PDFs. In this Section, we analyze whether the normalizing flow is able to reproduce the statistics of the actual time series. To this end, we compare the histograms of hourly prices in the realizations and scenarios as well as the leading statistical moments. In the scenario histograms, we scale the number of occurrences by the number of samples in order to match the realization histograms. 

\begin{figure}\centering
\includegraphics[width=\columnwidth]{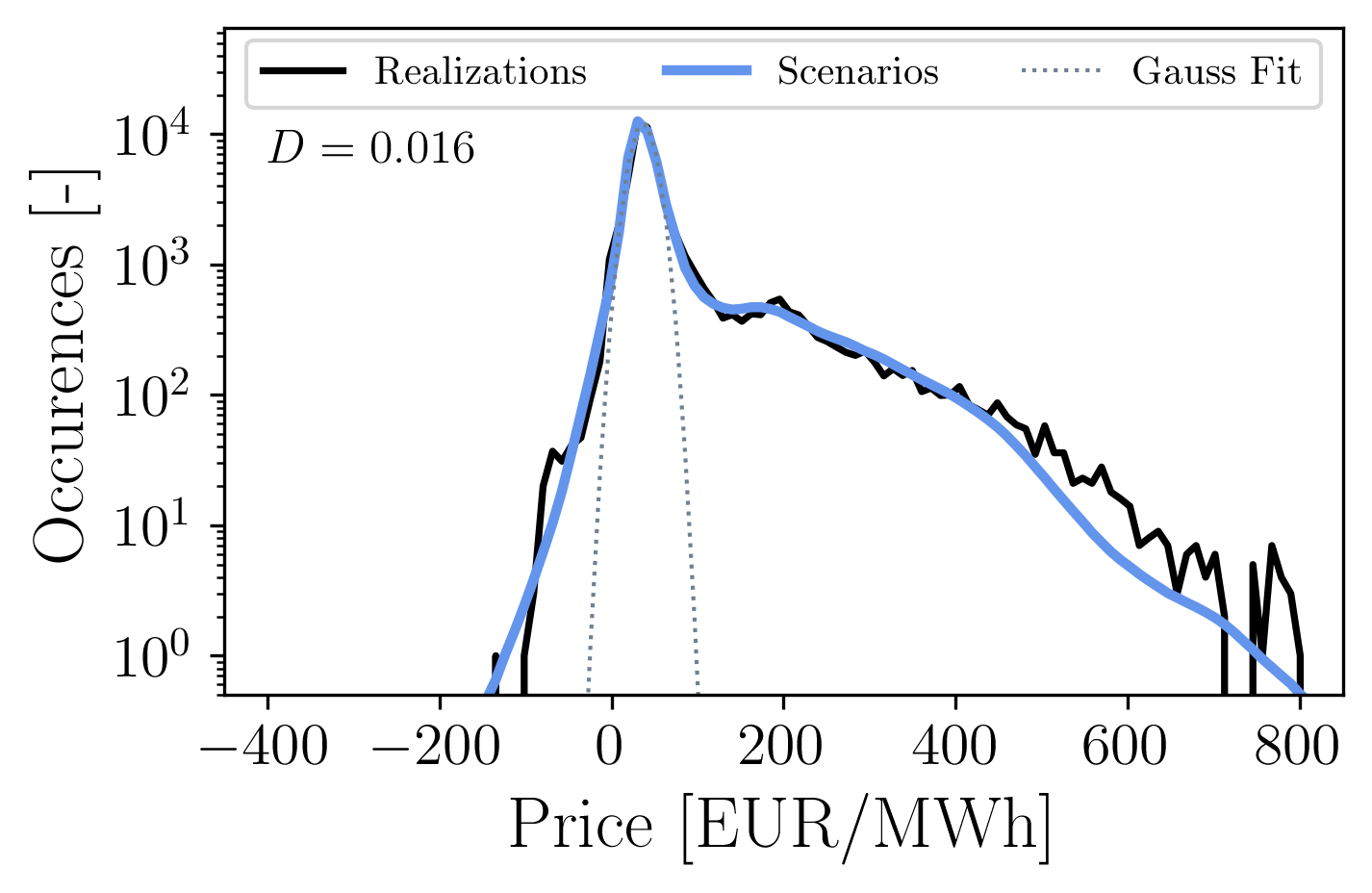}
    \caption{Histogram of prices of all generated scenarios compared to the histogram of the actual day-ahead price time series (``realizations''). Dotted line is a Gaussian fit onto the realizations histogram. The value $D$ gives the Kullback--Leibler divergence between scenario and realization histogram. Time series ranges from April 20, 2016, to December 31, 2022.}
    \label{fig:4}
\end{figure}

Figure~\ref{fig:4} shows the histogram for the entire period of analysis from April 20, 2016, to December 31, 2022. As motivated in our previous work \cite{cramer2022validation}, Figure~\ref{fig:4} shows the histogram in logarithmic scaling to allow for an analysis of the tails of the distribution. 
Overall, the scenario histogram matches the realizations histogram very well, which is also reflected in the similarity of the statistical moments listed in Table~\ref{tab:4}. 
The scenarios slightly underestimate the likelihood of high prices. This discrepancy is due to the stark increase in prices that limits the ability to adjust to changing market conditions. At the onset of the energy crisis, the normalizing flow underestimates the electricity prices as the price increase is not yet included in the training data. However, this period is rather short due to our retraining scheme such that we observe a very good overall agreement.

\begin{table*}\centering
\caption{Mean $\mu$, standard deviation $\sigma$, skewness $s$, and kurtosis $k$ of the day-ahead price time series (``realizations'') and of the scenarios generated by the normalizing flow.
We provide the normalized central moments for the entire time period under observation as well as separately for the time before and during the energy crisis.}
\begin{tabular}{llrrrr}
\toprule
                 & & $\mu$ & $\sigma$ & $s$  & $k$ \\ \midrule
2016-04-20 --  & realizations &  72.87 &  90.75 & 2.94 & 10.11 \\
2022-12-31    & scenarios    &  70.19 &  84.98 & 2.75 &  9.16 \\ \hline 
2016-04-20 -- & realizations &  40.38 &  23.63 & 1.29 &  7.17 \\
2021-09-30    & scenarios    &  39.17 &  22.83 & 1.54 &  8.57 \\ \hline 
2021-10-01 -- & realizations & 218.43 & 129.25 & 0.91 &  1.05 \\
2022-12-31    & scenarios    & 209.74 & 115.97 & 0.74 &  1.72 \\ \bottomrule
\end{tabular}
\label{tab:4}
\end{table*}

\begin{figure}\centering
\includegraphics[width=\columnwidth]{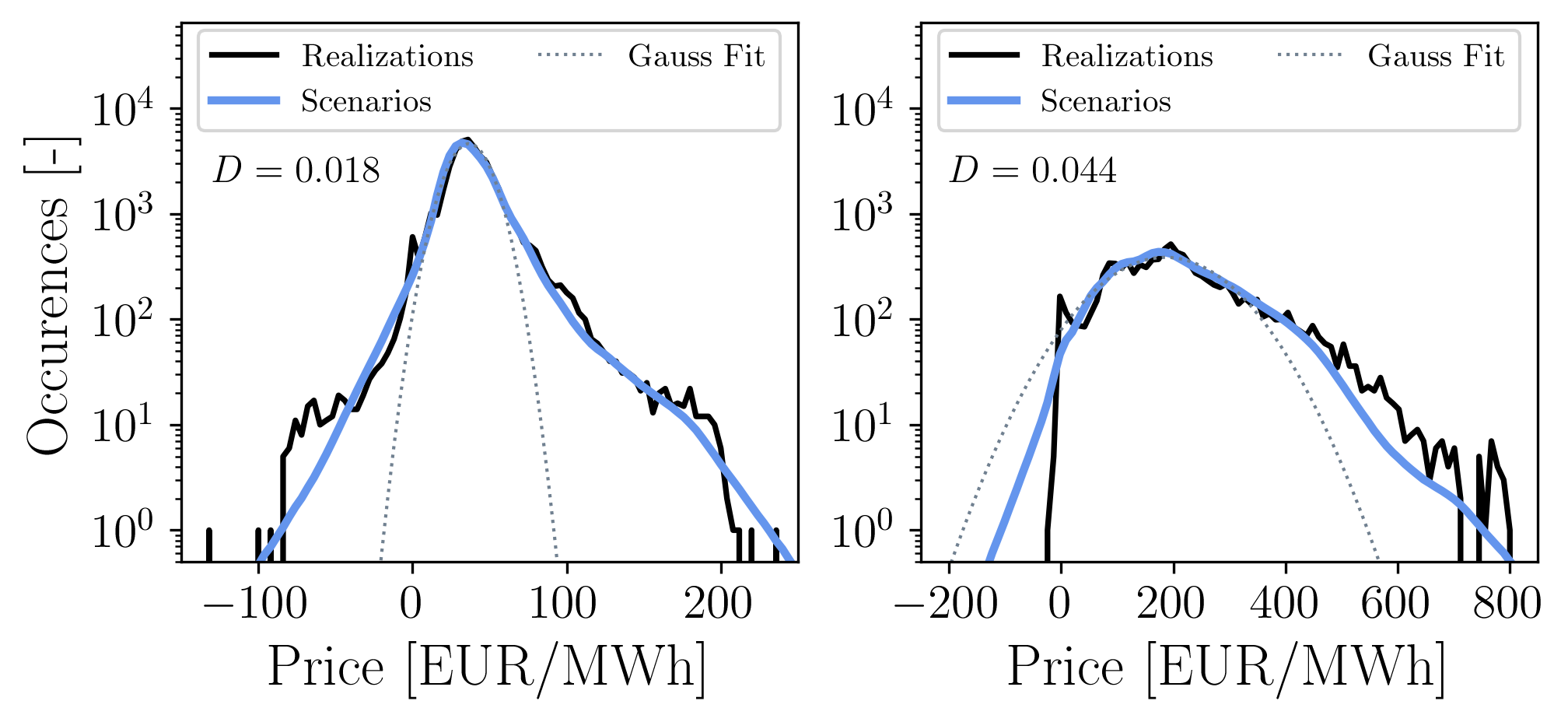}
    \caption{Histograms of prices of generated scenarios compared to histograms of the actual day-ahead price time series (``realizations''). The normalizing flow is trained on all available data at the given time. 
    The left side shows histograms for time series before October 1, 2021. The right side shows histograms for time series after October 1, 2021. Note the different scales on the x-axis. Dotted lines present Gaussian fits onto the realizations histograms. The value $D$ gives the Kullback--Leibler divergences between scenario and realization histograms.
    }
    \label{fig:5}
\end{figure}

We emphasize that the histograms have an unusual shape, which differs considerably from the histograms for the period 2015 to 2019 analyzed in \cite{han2022complexity}. This is a direct result of the overlay of distributions from different market regimes, i.e., before and during the energy crisis (Figure~\ref{fig:1}). 
For a more detailed analysis, we show separate histograms for the two market periods in Figure~\ref{fig:5}. The distribution of prices during the energy crisis vastly differs from the distribution before the crisis.
Notably, the scenarios show a good overall match to the realizations, demonstrating the normalizing flow's capability to learn and sample from complex non-Gaussian distributions.

The histograms show that negative electricity prices seldom occur after the onset of the energy crisis. However, the normalizing flow overestimates the occurrences and magnitudes of negative prices. 
The virtual absence of negative electricity prices has both economic and regulatory reasons~\cite{joas_energy_2022, joas_economics_2023}. In the German market, wind turbines and large solar PV installations receive subsidies (``Marktprämie'') that are given by the difference between a fixed reference value (``Anzulegender Wert'') and the average market price level.  
In a high-price market regime, the average market price level exceeds the reference value and the subsidies drop to zero. In such a case, wind and solar plants curtail generation to avoid negative prices. Hence, the price frequently decreases to zero or small positive values but rarely to negative values. Figure~\ref{fig:5} shows a small peak around zero but very few values below zero. 
As the proposed normalizing flow scheme is fully data-driven, this regulatory mechanism cannot be enforced explicitly. The scenarios fail to represent the respective features of the distribution and the scenario histogram is smoothed, missing the peak at zero and the sharp decrease below. 
We argue that this mismatch results from the change in regime being ahead of the adoption of training data.
Moreover, the training data for the normalizing flow after the onset of the energy crisis still includes data from previous years containing negative prices. 
Still, the results after the onset of the energy crisis show limitations of the normalizing flow.

\begin{figure}\centering
\includegraphics[width=0.49\columnwidth]{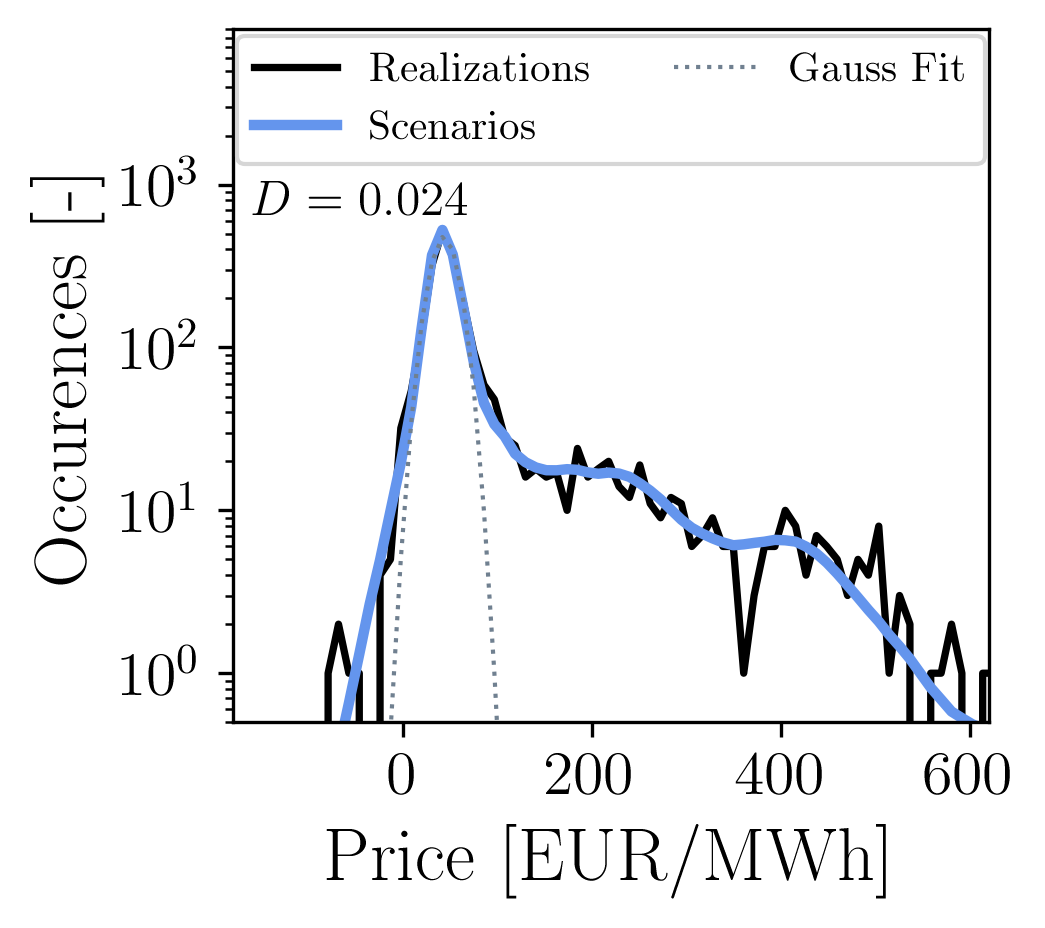}
    \includegraphics[width=0.49\columnwidth]{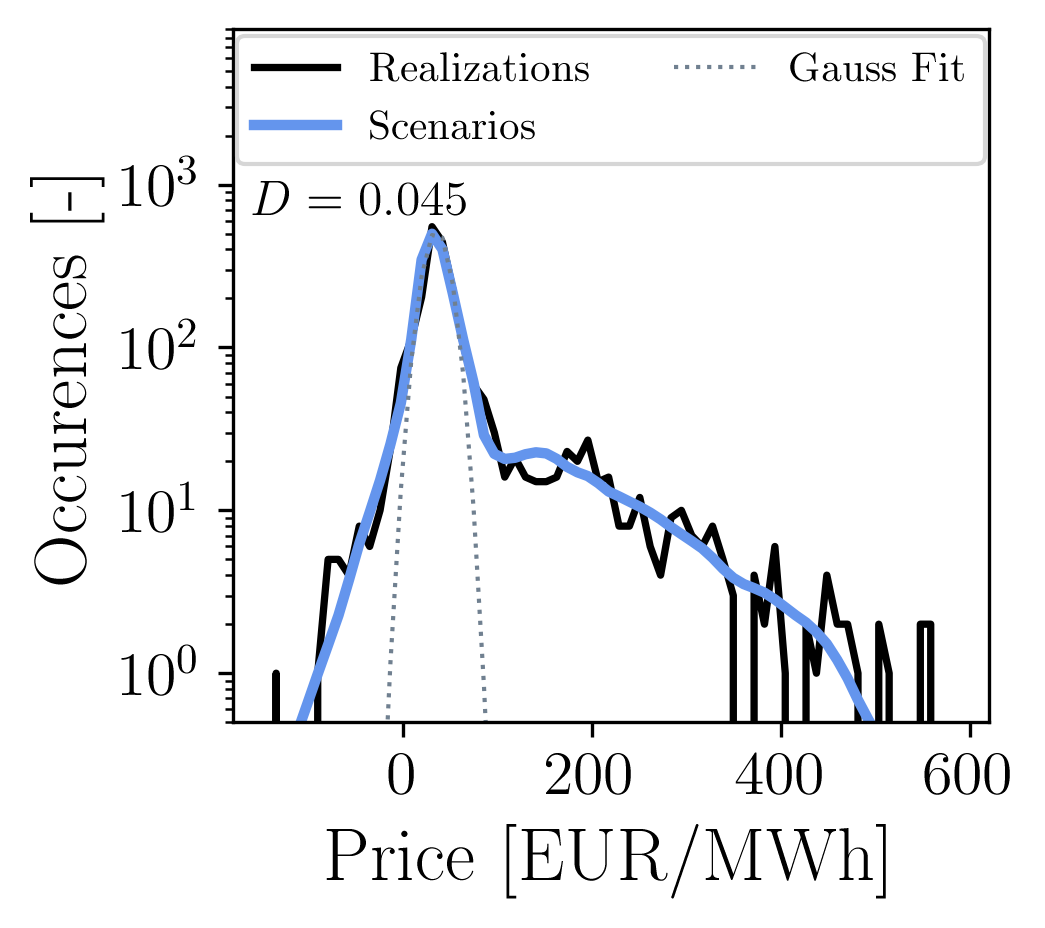}
    \caption{Marginal price histogram of generated scenarios vs. realizations at 06:00 (left) and 12:00 (right).}
    \label{fig:6}
\end{figure}

Beyond the full distributions, we emphasize that the normalizing flow also reproduces the \emph{marginal} distributions.
Figure~\ref{fig:6} shows the histograms for two hourly windows starting at 06:00 and 12:00. The probability for high prices is higher at 06:00, while the probability for low or negative prices is higher at 12:00.
Again, this is well explainable through a typical solar profile and the merit order effect.
The generated scenarios reflect this behavior and produce different distributions for different hours of the day.

\subsection{Forecasting performance}

\begin{figure}\centering
\includegraphics[width=\columnwidth]{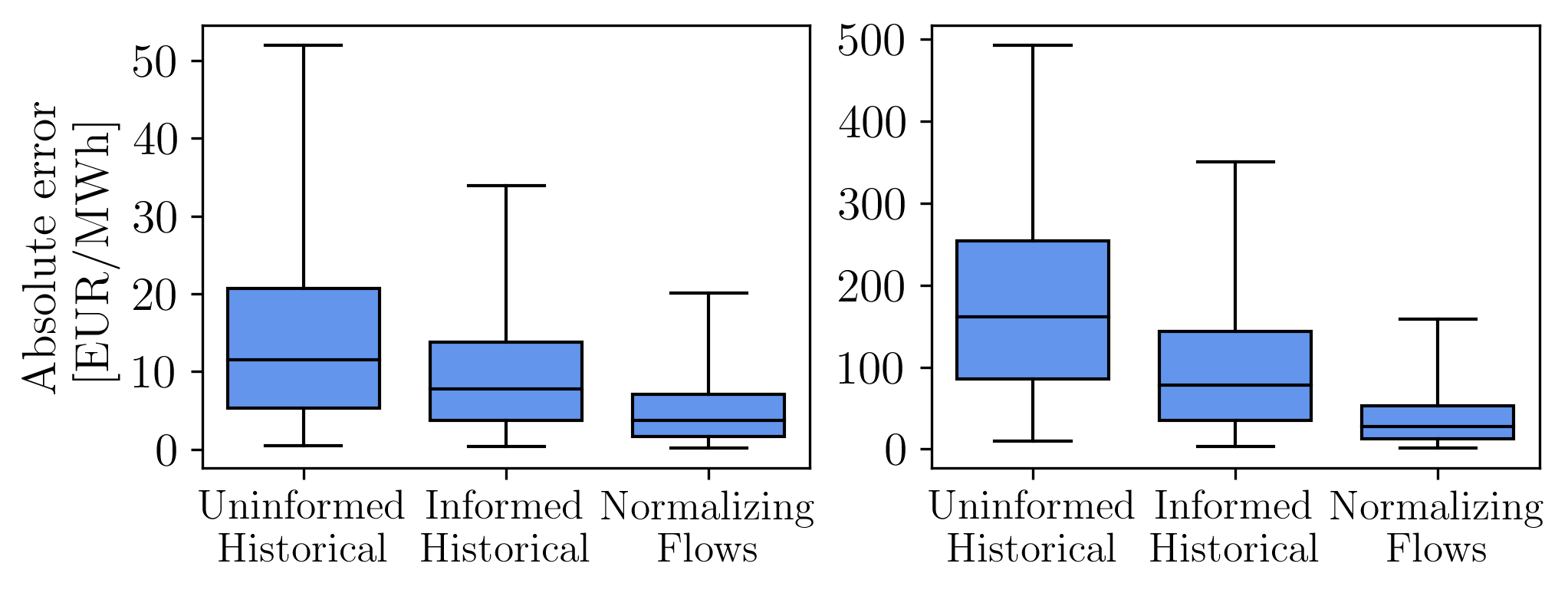}
    \includegraphics[width=\columnwidth]{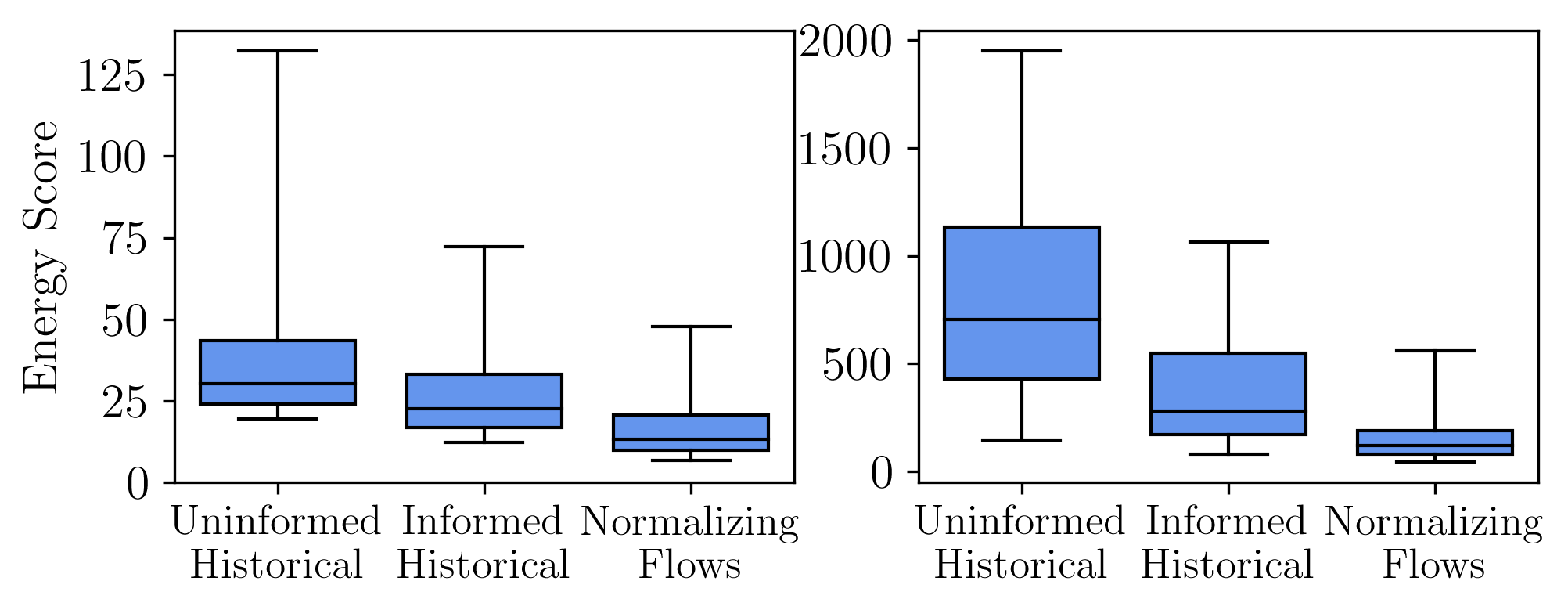}
    \includegraphics[width=\columnwidth]{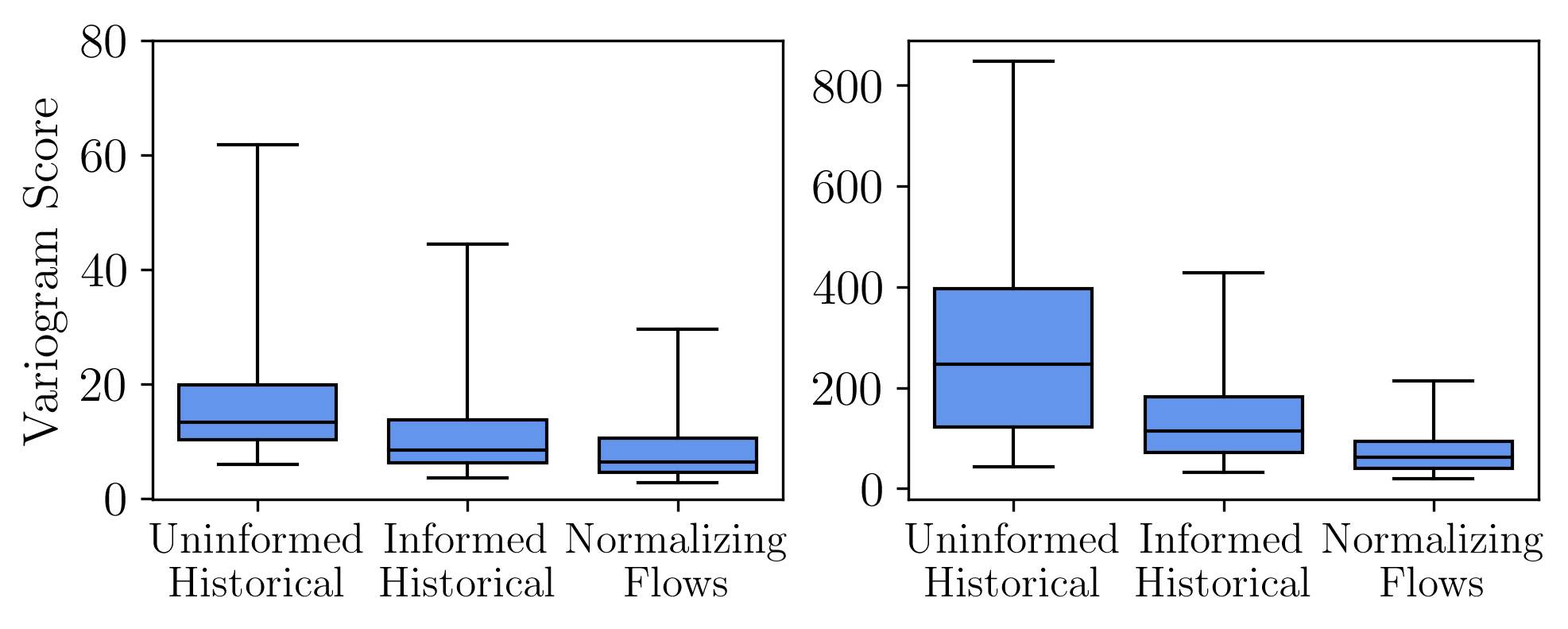}
    \caption{
    Performance of the normalizing flow in comparison to the benchmark models. 
    The upper plots compare the mean absolute error (MAE) for 2019 (left) and 2022 (right). 
    The center plots compare the energy score for 2019 (left) and 2022 (right). 
    The bottom plots compare the variogram score for 2019 (left) and 2022 (right). 
    The black vertical bar indicates the sample median. The boxes indicate the ranges between 75\% and 25\%, and the whiskers indicate the range between 97.5\% and 2.5\%.} 
\label{fig:7}
\end{figure}

We provide a quantitative assessment of the performance of the normalizing flow in reference to the two benchmark scenario generation methods.
First, we show the mean absolute error (MAE), i.e.,~the MAE of the hourly mean values of the generated scenarios.
We emphasize that the MAE is designed to evaluate point forecasts.
Thus, our MAE analysis is limited to the mean of the generated scenarios.
Results are provided in Figure~\ref{fig:7} in comparison to the two benchmark models introduced in Section~\ref{sec:benchmark}.
We find that the normalizing flow strongly outperforms the two benchmark models in terms of the MAE.
In particular, under shifting market conditions such as in 2022, the normalizing flow approach with retraining holds up well.
For the period from 2019-01-30 to 2020-02-08, we find an MAE of $3.88$\,EUR/MWh for the mean value of the normalizing flow scenarios. 
This value is comparable to our recent results using LSTM models \cite{trebbien2023probabilistic}, reporting state-of-the-art performance with an MAE of $3.73$\,EUR/MWh for the year 2019.
For the period between the years 2019--2022, our previous work \cite{trebbien2023probabilistic} finds an MAE of $11.92$\,EUR/MWh. 
Again, the normalizing flow yields competitive results with an MAE of $11.11 \pm 0.56$\,EUR/MWh over the entire period of 2016--2022 (cf. Table~\ref{tab:3}). 

Considering the different time periods, the results from \cite{trebbien2023probabilistic} are slightly better.
The time before the energy crisis, which generally leads to a lower MAE due to the lower absolute price values, is more strongly represented in the full data set used for the normalizing flow training.
Nevertheless, we find that the normalizing flow is generally competitive even in terms of the MAE despite not being designed to produce point forecasts.

In contrast to the MAE, there are metrics to specifically evaluate the quality of probabilistic forecasts.
As in our previous work on intraday price forecasting \cite{CRAMER2023_deltaID3}, we use the energy score (ES) and the variogram score (VS). 
The energy score \cite{gneiting_strictly_2007,pinson_evaluating_2012} is defined as
\begin{align} 
    \text{ES} = \frac{1}{N} \sum_{s=1}^{N} \left \| \mathbf{\lambda} - \mathbf{\hat{\lambda}_s} \right \|_2 - \frac{1}{2N^2} \sum_{s=1}^{N} \sum_{s'=1}^{N} \left \| \mathbf{\hat{\lambda}_s} - \mathbf{\hat{\lambda}_{s'}} \right \|_2.
    \label{eq:def-ES}
\end{align}
Here, $\lambda$ is the 24-dimensional realized price profile for a given day and $\hat{\lambda}_s$ is the price profile per scenario $s$.
The operator $\left \| \cdot \right \|_2$ denotes the Euclidean norm and $N$ is the number of scenarios used to compute the energy score. 
The first term on the right side of Equation~\eqref{eq:def-ES} measures the distance between the scenarios and the realization. The second term measures the diversity of the samples. 
The VS~\cite{scheuerer_variogram-based_2015} quantifies whether the forecasts correctly describe the correlations between the individual time steps. It is defined as
\begin{align} 
    \text{VS} = \frac{1}{N} \sum_{t=1}^{T} \sum_{t'=1}^{T} \left ( |\lambda_t - \lambda_{t'}|^{\gamma} - \frac{1}{N} \sum_{s=1}^{N} | \hat{\lambda}_{t,s} - \hat{\lambda}_{t',s} |^{\gamma} \right )^2.
\end{align}
The parameter $\gamma$ is referred to as variogram order and is typically set to $\gamma = 0.5$~\cite{scheuerer_variogram-based_2015}. Both ES and VS scores are negatively oriented, i.e., a lower score indicates a better result.
Similarly, $N$ is the number of scenarios and $T$ is the number of time steps within each scenario. 

Figure~\ref{fig:7} shows box plots of the MAE, ES, and VS distributions in comparison to the two benchmark models, before and after the beginning of the energy crisis.
The results show that the normalizing flow yields substantially lower values for both scores, thus indicating a much better agreement with the realizations.
Furthermore, the normalizing flow consistently outperforms the benchmark methods for both periods and even increases its advantage in 2022.
Notably, the absolute values of the MAE, ES, and VS increase by about a factor of ten with the onset of the energy crisis.
This increase is expected as the absolute prices increase by a similar factor.
We made the same observation in our previous work on price forecasting using LSTM models \cite{trebbien2023thesis}.

In summary, the analysis in this Section shows how conditional normalizing flows can generate high-quality scenarios of day-ahead electricity prices. The normalizing flow generates realistic scenarios, and the adaptive retraining of the normalizing flow produces high-quality results throughout the transition of market regimes.

\subsection{Correlations}
\begin{figure}\centering
    \includegraphics[width=\columnwidth]{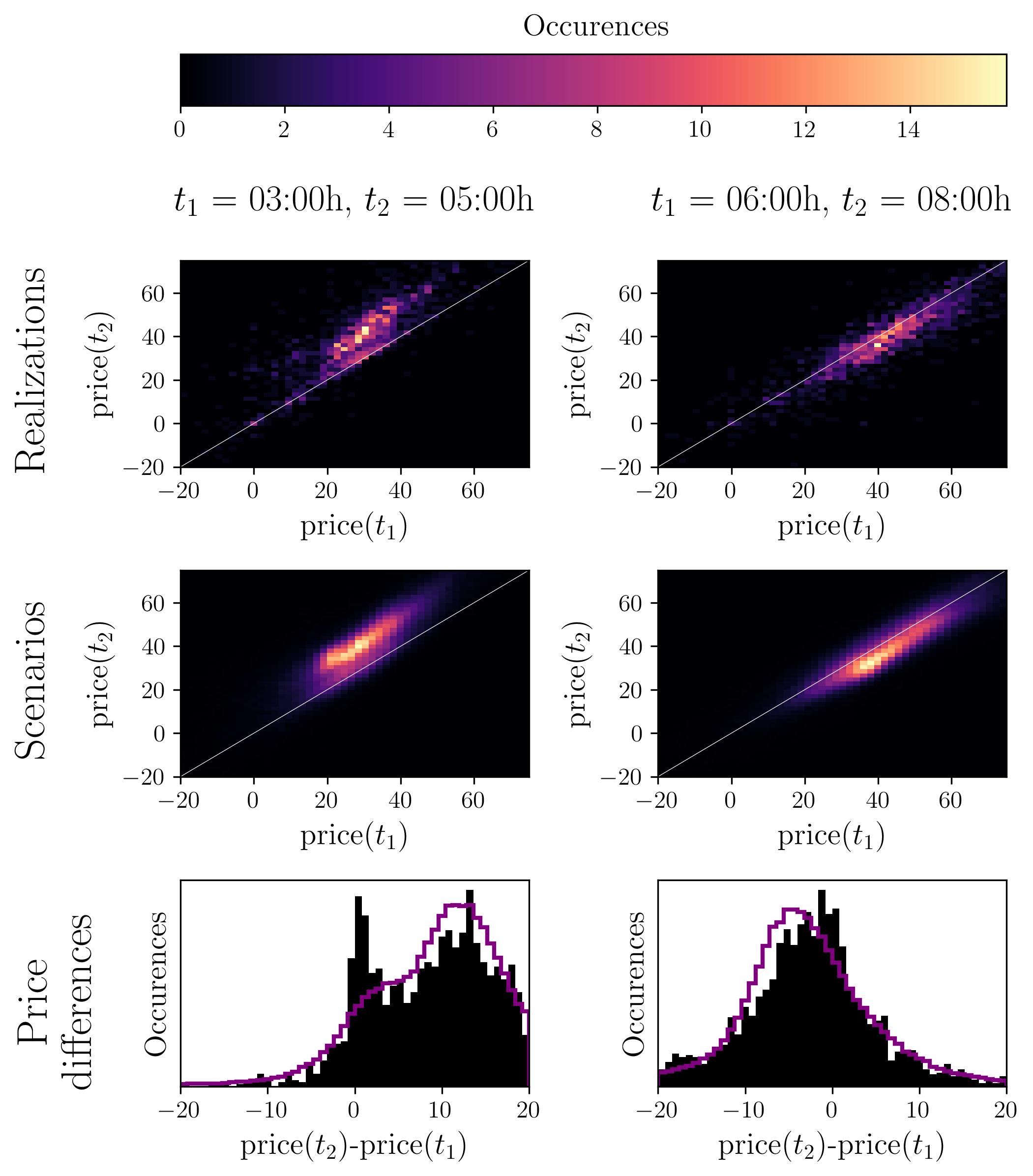}
    \caption{
    Reproduction of correlations in the price time series.
    We investigate the joint probability distribution for two points in time (left: $t_1$\,=\,03:00 and $t_2$\,=\,05:00, right: $t_1$\,=\,06:00 to $t_2$\,=\,08:00).
    We show the histograms of the realizations (top panels) and scenarios (center panels). 
    The lower panel shows statistics of the increments $\Delta = \mbox{price}(t_2) - \mbox{price}(t_1)$. The black bars represent the increment histograms of the true realizations, the purple lines represent the increment histograms of the generated scenarios.
    } 
    \label{fig:8}
\end{figure}

An important advantage of multivariate scenario forecasting is that each scenario is intrinsically consistent, i.e., every generated scenario reflects correlations present in the actual price time series.
Mathematically speaking, the normalizing flow learns the distribution of a random vector $X$ describing the prices of an entire day instead of an individual hour.

We test the capability of the model by fixing two points in time, $t_1$ and $t_2$, and comparing the \emph{joint} probability distribution of the respective prices. Figure~\ref{fig:8} shows the histograms of the occurrences for the two respective times.
In the early morning, prices increase from $t_1$\,=\,03:00 to $t_2$\,=\,05:00 in a characteristic way (see~Figure~\ref{fig:3}).
Hence, the joint PDF is concentrated above the bisector.
Later, between $t_1$\,=\,06:00 and $t_2$\,=\,08:00, the prices mostly decrease and the joint PDF is concentrated slightly below the bisector.
In both cases, Figure~\ref{fig:8} shows that the normalizing flow reproduces the joint PDF aptly, and thus successfully learns the correlations between the different points in time. 

For a more detailed analysis, we consider the price increments $\Delta = \mbox{price}(t_2) - \mbox{price}(t_1)$ and compute their histogram (Fig.~\ref{fig:8} bottom).
Overall, we find a good agreement of the scenarios and realizations in terms of the increment statistics.
The increment histograms of the scenarios (purple lines) reproduce the overall shape of the increment histograms of the realizations (black bars).
However, in both examples, the actual realizations show a sharp peak at an increment of $\Delta \approx 0$, which is not reproduced.
This peak results from complex regulatory aspects of the market.
For instance, as discussed in Section~\ref{sec:stats_nfresults}, the regulation of renewable subsidies leads to an increased likeliness of a price of 0\,EUR/MWh or slightly above.
Hence, there is an increased likelihood that the price stays at a fixed value for several hours leading to an increment of $\Delta \approx 0$\,EUR/MWh.
The normalizing flow does not learn this characteristic such that the increment distribution is smoothed compared to the actual data.
Again, we expect this model behavior to change with the inclusion of more training data from later periods.
Furthermore, excluding of data from earlier periods where negative prices were more prevalent may further improve the results.

\subsection{Errors and uncertainties}

\begin{figure}\centering
\includegraphics[width=\columnwidth]{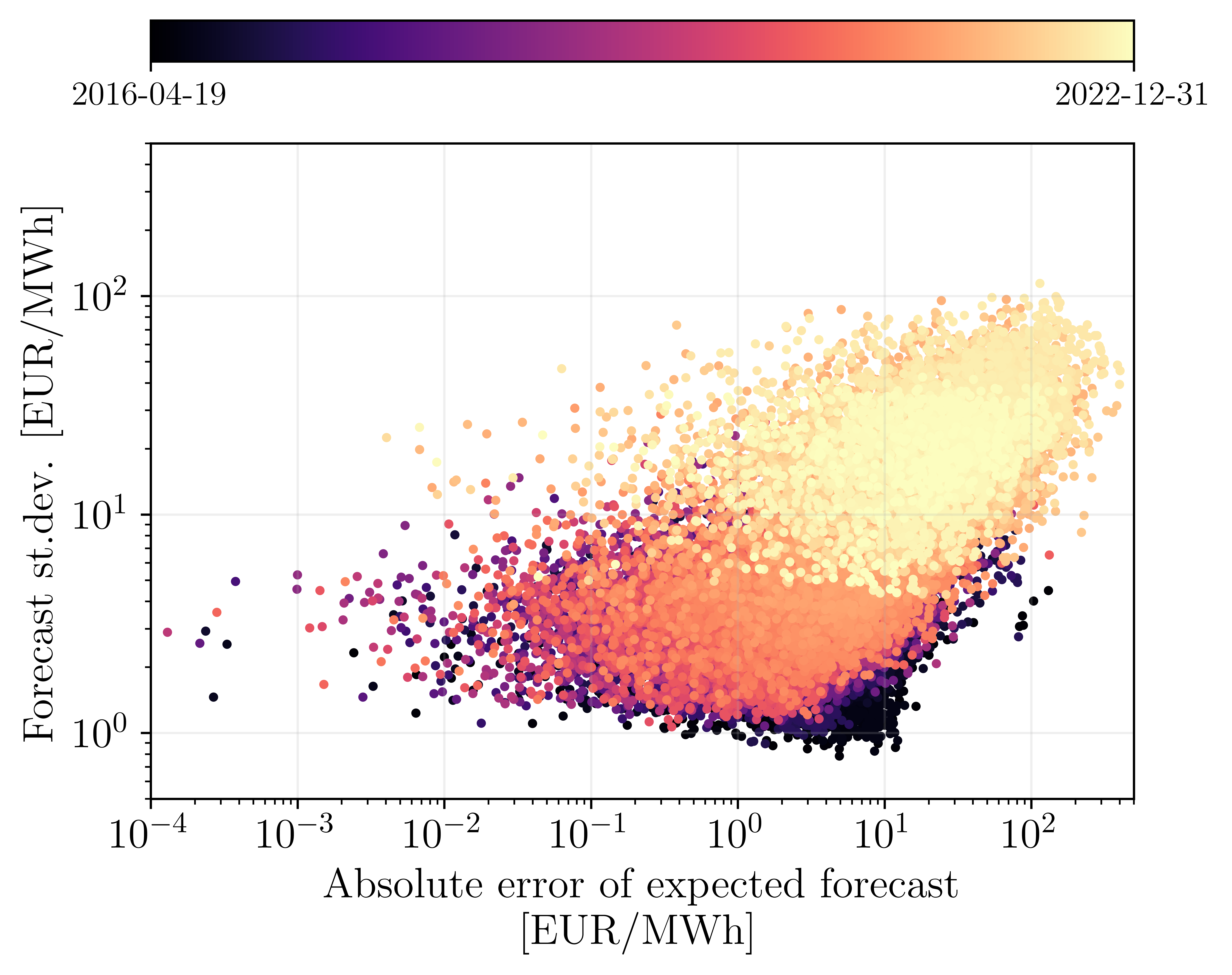}
    \caption{Standard deviation of hourly forecast distribution against absolute error of the expected forecast. Each dot represents one hour. Color represents the date according to the color bar.}
    \label{fig:9}
\end{figure}

Quantification of forecast uncertainty is of high importance in many applications.
We consequently study whether the normalizing flow can provide a measure of confidence for its forecasts.
In particular, we examine the following question:
If the scenario mean has a high error for a particular hour, did the model express uncertainty about the outcome?
In Figure~\ref{fig:9}, we compare the standard deviation of the hourly forecast distribution to the MAE of the mean value of the generated scenarios. 
The scatter plot reveals that there is indeed a correlation between the MAE of the expected forecast and the forecast standard deviation, i.e.,~events with a high MAE of the expected forecast but a low forecast standard deviation rarely occur.
Note that the correlation between MAE of the expected forecast and its attributed standard deviation have no strict correlation and, there are instances with low standard deviation and relatively high forecast errors. 
Still, there appears to be a lower bound of the standard deviation for higher forecast errors. Thus, this lower bound should be the criteria for the quality assessment of the scenario forecast.
In summary, the normalizing flow provides information on how trustworthy the predictions are as low-confidence forecasts come with a high standard deviation. 
We observe this type of uncertainty representation for most test data. However, there is a variance in the assigned level of the uncertainty.

Similar to Figure~\ref{fig:7}, the change in behavior over time in Figure~\ref{fig:9} shows increasing absolute errors and standard deviations for later periods.
The observed lower bound of the forecast standard deviation increases over time with the change of the market regime.
This behavior is consistent with our expectation of adjusting towards the high-price regime with higher variance after the onset of the energy crisis. 
Figure~\ref{fig:1} shows that after the onset of the energy crisis both the absolute electricity prices and their fluctuation increased drastically.
Thus, increased absolute errors, energy scores, and variogram scores are expected as a result of larger absolute values of the data. 
Similarly, the variance predicted for the outcome also increases as the fluctuations increase.
The results in Figure~\ref{fig:9} show that with the progression of time both the absolute error and the forecast standard deviation, i.e., the uncertainty estimate, increase in the same order of magnitude. 
In summary, the progression shown in Figure~\ref{fig:9} confirms our observation that the normalizing flow with periodic retraining adapts to changing market conditions and also adjusts its estimate of the uncertainty of the forecasts.

\section{Conclusion and Outlook}
\label{sec:conclusion}

We present a multivariate probabilistic forecasting approach for day-ahead electricity prices based on normalizing flows.
Our normalizing flow implementation incorporates relevant feature information to learn the conditional multivariate probability distribution of the vector of day-ahead electricity prices.
We train our model via direct log-likelihood maximization to achieve mathematically consistent and efficient training. 
The trained model allows for sampling day-specific scenarios of electricity price time series that are intrinsically consistent and match the fundamental market structure of the day-ahead bidding market of the EPEX spot markets by generating full 24-hour scenarios.

Our analysis shows that the normalizing flow yields high-quality scenarios with a good representation of the actual price realization and informative uncertainty quantification that indicates the reliability of the forecasts in a quantitative way.
The conditional normalizing flow significantly outperforms uninformed historical sampling and KNN-based selection of historical scenarios.
Still, our analysis shows that the normalizing flow has some limitations w.r.t. learning effects stemming from regulatory standards in the markets. 
This aspect may be addressed in future research, e.g., by including regulatory aspects directly. In particular, the subsidy reference price could be included as a further conditional input.

We propose a periodic retraining scheme to continuously adapt the normalizing flow to the changes in market regimes such as the onset of the energy crisis in 2021.
With brief delays, the normalizing flow adapts to the changing markets and generates high-quality scenarios.
This retraining scheme could prove useful for analyzing and modeling other strongly non-stationary time series.

\section*{Acknowledgments}

The authors gratefully acknowledge the computing time granted through JARA on the supercomputer {JURECA}~\cite{JURECA} at Forschungszentrum J{\"u}lich.
E.C. gratefully acknowledges the financial support of the Kopernikus project SynErgie 3 by the Federal Ministry of Education and Research (BMBF) and the project supervision by the project management organization Projektträger Jülich.
D.W., H.H., J.T., and M.D. received funding from the Helmholtz Association of German Research Centers.

    \bibliographystyle{unsrt}
\renewcommand{\refname}{Bibliography}

  \bibliography{references}

\end{document}